\documentclass{article} 
\usepackage{fullpage}
\usepackage{soul,color}
\usepackage{tabularx}
\usepackage{booktabs}
\usepackage{multirow}
\usepackage{graphicx}
\usepackage{subcaption}
\usepackage{soul}
\usepackage[dvipsnames]{xcolor}
\usepackage[table]{xcolor}

\usepackage{balance} 
\usepackage{url}
\usepackage[affil-it]{authblk}

\AtBeginDocument{%
  }

\begin{document}

\title{Investigating the Generalizability of ECG Noise Detection Across Diverse Data Sources}
\date{}
\author{Sharmad Kalpande, Nilesh Kumar Sahu, Haroon R. Lone}
\affil{Indian Institute of Science Education and Research Bhopal, India}

\maketitle

\begin{abstract}
Electrocardiograms (ECGs) are vital for monitoring cardiac health, enabling the assessment of heart rate variability (HRV), detection of arrhythmias, and diagnosis of cardiovascular conditions. However, ECG signals recorded from wearable devices are frequently corrupted by noise artifacts, particularly those arising from motion and large muscle activity, which distort R-peaks and the QRS complex. These distortions hinder reliable HRV analysis and increase the risk of clinical misinterpretation.

Existing studies on ECG noise detection typically evaluate performance on a single dataset, limiting insight into the generalizability of such methods across diverse sensors and recording conditions. In this work, we propose an HRV-based machine learning approach to detect noisy ECG segments and evaluate its generalizability using cross-dataset experiments on four datasets collected in both controlled and uncontrolled settings. Our method achieves over 90\% average accuracy and an AUPRC exceeding 90\%, even on previously unseen datasets—demonstrating robust performance across heterogeneous data sources. To support reproducibility and further research, we also release a curated and labeled ECG dataset annotated for noise artifacts\footnote{\textbf{This is the author's accepted manuscript of a paper accepted for publication in ACM ISWC'25. 
The final published version is available via the ACM Digital Library at \url{<https://dl.acm.org/doi/10.1145/3715071.3750402>}.}}.
\end{abstract}

\section{Introduction}

Electrocardiogram (ECG) monitoring is essential for detecting and managing life-threatening cardiac conditions. While standard 12-lead ECG systems offer detailed cardiac insights in clinical settings, their bulkiness and immobility limit their utility for continuous, real-world monitoring \cite{corrado2010recommendations}. Recent advancements in wearable technology have enabled the development of compact ECG devices using three \cite{kristensen2016use}, two \cite{moody1984bih}, or even a single electrode \cite{nemcova2020brno}, supporting real-time, ambulatory heart monitoring. However, ECG signals captured by wearable sensors are highly susceptible to noise artifacts \cite{chatterjee2020review}, which compromise signal quality and clinical interpretability.

\begin{figure*}[!t]
    \centering
    \begin{subfigure}[b]{0.24\textwidth}
        \centering
        \includegraphics[width=\textwidth]{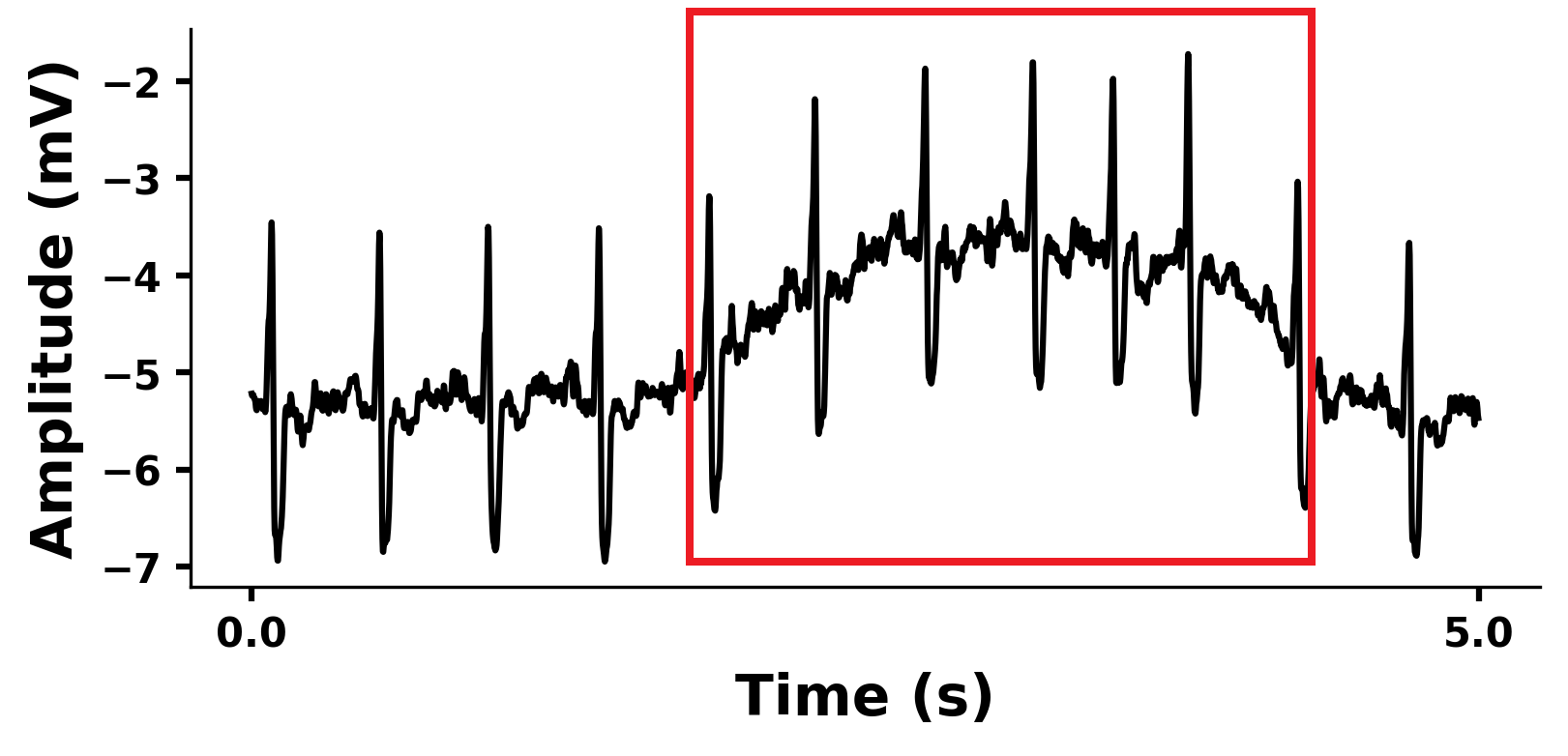}
        \caption{Baseline Wander}
        \label{fig:Baseline Wander}
    \end{subfigure}
    \hfill
    \begin{subfigure}[b]{0.24\textwidth}
        \centering
        \includegraphics[width=\textwidth]{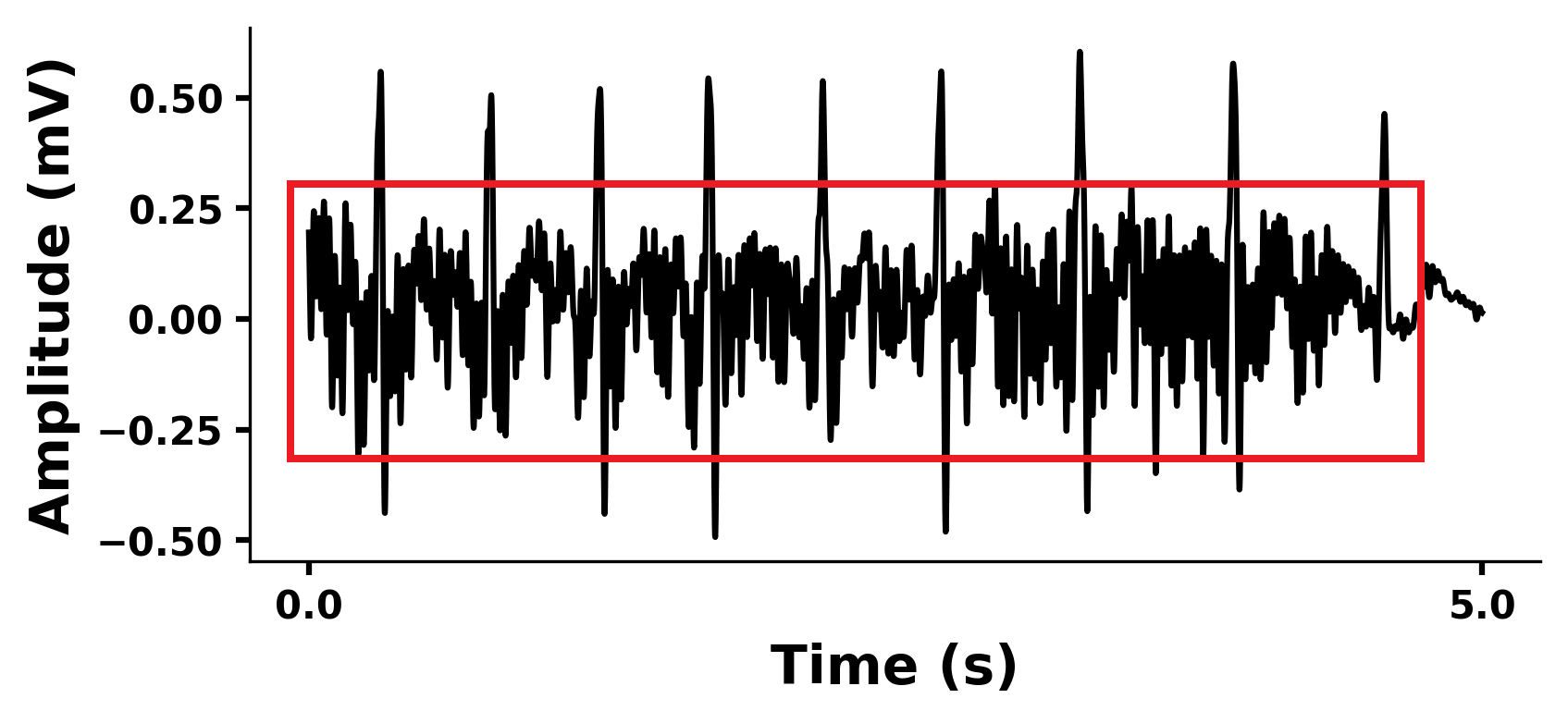}
        \caption{Powerline Interference}
        \label{fig:Powerline Interference}
    \end{subfigure}
    \hfill
    \begin{subfigure}[b]{0.24\textwidth}
        \centering
        \includegraphics[width=\textwidth]{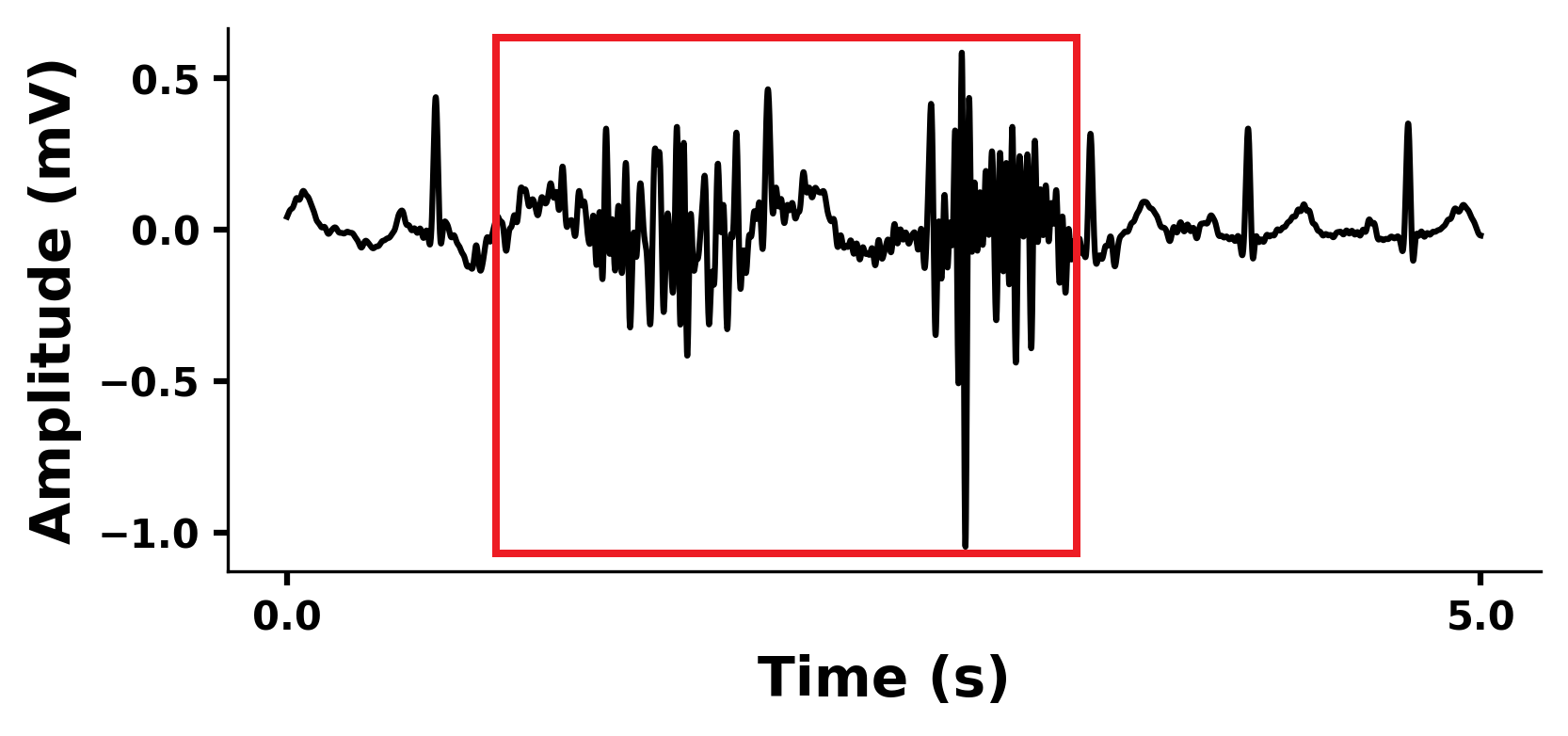}
        \caption{Muscle Artifact}
        \label{fig:muscle-artifact}
    \end{subfigure}
    \hfill
    \begin{subfigure}[b]{0.24\textwidth}
        \centering
        \includegraphics[width=\textwidth]{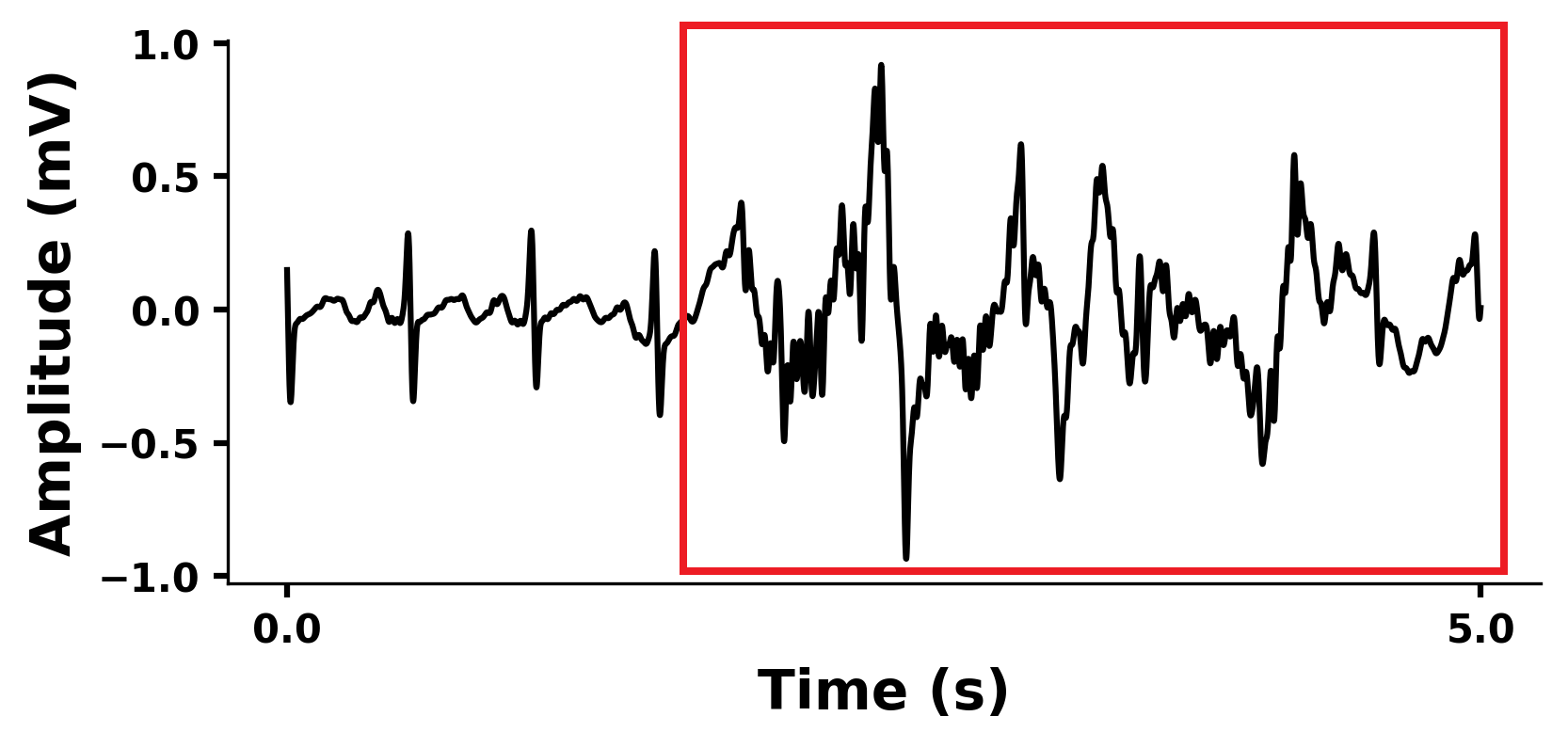}
        \caption{Motion Artifact}
        \label{fig:Motion Artifact}
    \end{subfigure}

    \caption{Annotated noisy signatures: (a) Baseline Wander, (b) Powerline Interference, (c) Muscle, and (d) Motion Artifact.}
    \label{fig:ECG_artifacts}
\end{figure*}

Noisy ECG signals typically falls into four categories: (i) \textit{Baseline wander}—low-frequency drifts due to respiration and electrode-skin interface (see Fig. \ref{fig:Baseline Wander}); (ii) \textit{Power line interference}—sharp oscillations from electrical sources (see Fig. \ref{fig:Powerline Interference}); (iii) \textit{Muscle artifacts}—high-frequency bursts from tremors or movement (see Fig. \ref{fig:muscle-artifact}); and (iv) \textit{Motion artifacts}—abrupt, irregular signal shifts due to body motion (see Fig. \ref{fig:Motion Artifact}). While conventional filters can mitigate baseline wander and minor interferences, motion and large muscle artifacts are particularly problematic. These distort the QRS complex and R-peaks, impeding both morphological analysis and RR interval extraction \cite{huang2018detection,sahu2024wearable}. Our work focuses on motion and muscle artifacts.

Recent Artificial Intelligence (AI) research has applied machine learning to ECG signals for several applications~\cite{abubaker2022detection,sahu2024wearable}. However, many approaches rely on manual noise removal, limiting scalability \cite{sahu2024wearable}. Attempts to automate noise detection have emerged \cite{dua2022automatic,yoon2019deep}, yet they often lack generalizability. Variations in movement type, sensor hardware (e.g., Shimmer, Bittium Faros 180, Holter monitors), skin-electrode contact, sampling rates, and recording contexts (controlled vs. uncontrolled) introduce significant heterogeneity, making robust, cross-context noise detection challenging.

We propose an AI-based method for detecting noisy ECG segments that generalizes across multiple sensor platforms and recording environments. We define a segment as noisy if it contains motion or large muscle artifacts that obscure R-peaks. Heart rate variability (HRV) features are extracted from each segment, and machine learning models are trained to classify segments as noisy or clean. Among the models tested, Random Forests (RF) demonstrated the best within-dataset performance and were selected for cross-dataset evaluation.

We assess our approach using four diverse ECG datasets. To our knowledge, this is the first study to evaluate cross-dataset generalizability in ECG noise detection. Our model achieves an average accuracy of 91.1\% in cross-dataset tests and 93.6\% when trained on combined datasets, indicating strong generalization performance. 

\begin{itemize}
    \item[$\bullet$] We present a novel AI-based approach for detecting noisy ECG segments, particularly those affected by motion and large muscle artifacts. Moreover, we systematically assess the \textbf{generalizability} of our approach through cross-dataset and cross-combined dataset experiments. 

    \item We release a curated, annotated ECG dataset\footnote{\url{https://osf.io/dtbv8/}} labeled for motion and muscle artifacts, collected under semi-controlled conditions.
 
\end{itemize}

\section{Related Work}

Traditional signal processing techniques, such as Empirical Mode Decomposition (EMD), segment ECG signals and identify noise via statistical thresholding \cite{satija2017automated,lee2011automatic,kumar2020detection}. Building on this, signal quality indices (SQIs) have been widely used to classify ECG segments as noisy or clean \cite{satija2018review,zhao2018sqi,rahman2022robustness}, with tools like NeuroKit2 relying on rule-based thresholds over sliding windows. However, these thresholds are typically dataset-specific and fail to generalize to unseen data \cite{rahman2022robustness}. Heuristic SQI methods also struggle with signal variability, producing inconsistent results and motivating the use of machine learning (ML) approaches \cite{satija2018review}.

ML-based methods often split ECG signals into short windows \cite{holgado2023characterization,li2014machine,vijayakumar2022ecg} and apply feature extraction followed by classification \cite{holgado2023characterization,vijayakumar2022ecg}. While these models outperform fixed-threshold SQI techniques, their reliance on hand-crafted features and limited, homogeneous training datasets reduces their robustness.

Deep learning (DL) approaches address some of these limitations by learning features directly from data \cite{dua2022automatic,vijayakumar2022ecg,holgado2023characterization,yoon2019deep}. Yet, class imbalance remains a challenge, often leading to the exclusion of clean segments \cite{vijayakumar2022ecg}. Specific models illustrate these limits: a 1D convolutional autoencoder for multi-class noise fails to generalize across datasets \cite{dua2022automatic}; a two-branch model based on peak count achieves perfect accuracy, but only on an undersampled dataset \cite{li2014machine}; and a 1D CNN combining time- and frequency-domain features is tested on a very small subset \cite{vijayakumar2022ecg}. Even large-scale studies, such as one benchmarking six DL models on six PhysioNet datasets—reporting up to 97.31\% accuracy with VGG and 95.07\% with ResNet—use synthetically induced noise and omit cross-dataset testing, limiting their real-world applicability \cite{rahman2024robustness}. Table\ref{tab:related-works-new} summarizes key studies in this space. 

Despite growing interest, the generalizability of ECG noise detection remains largely unaddressed. Most prior work trains and validates on a single dataset under controlled conditions, overlooking performance across diverse real-world sources and mostly evaluated on dataset with induced noise. While cross-dataset generalization has been explored in related domains such as stress detection \cite{prajod2022generalizability,benchekroun2023cross,prajod2024stressor} and arrhythmia classification \cite{huang2023generalization}, ECG noise detection has yet to benefit from such advances. To our knowledge, no existing study explicitly focuses on generalizing noise detection across multiple ECG datasets.

Our work distinguishes itself by employing an HRV based machine learning approach that generalizes across diverse datasets, encompassing both simulated and real-world noise conditions.

\begin{table*}[]
    \centering
    \scriptsize
    \renewcommand{\arraystretch}{1.5} 
    \setlength{\tabcolsep}{8pt}      
    \caption{Summary of ECG noise detection studies}
    \label{tab:related-works-new}
    \resizebox{\textwidth}{!}{%
      \begin{tabular}{>{\raggedright\arraybackslash}p{0.3cm} 
                      >{\raggedright\arraybackslash}p{4.2cm} 
                      >{\raggedright\arraybackslash}p{5.2cm} 
                      >{\raggedright\arraybackslash}p{5.2cm}}
        \hline
        \textbf{Ref.} & \textbf{Dataset(s)} & \textbf{Goal} & \textbf{Drawback(s)} \\
        \hline \hline
        \cite{satija2017automated} & MIT-BIHA, PCICC2011, PTBDECG, MIT-BIHSTC, and Fantasia & Developed ECG classification (noisy, clean) method & Relies on non-generalizable fixed thresholds. \\

        \cite{holgado2023characterization} & Private dataset & Characterized noise in long-term ECG signals. & Small dataset and low F1 (0.77) score\\
        
        \cite{falk2014ms} & Synthetic data, Garment ECG data, Physionet Challenge Signal Quality, MIT-BIH Arrhythmia &Developed a new ECG quality index called MS-QI.& Does not classify the ECG as noisy or clean. \\
        
        \cite{dua2022automatic} & BUTQDB and Private & Developed a noise detection method without class labels. & Did not test generalizability and resulted in low accuracy.\\
        
        \cite{yoon2019deep} & Private dataset & Developed a deep-learning based noise detection method. & Small dataset resulted in a limited evaluation. \\
        
        \cite{li2014machine} & Simulated and MITDB & Performed ECG signal quality classification & Simulated dataset and data is dropped during segmentation.\\
        
        \cite{vijayakumar2022ecg} & MITNSTD & Performed ECG signal classification & Dropped clean signal data to handle class balance. \\

        \cite{rahman2024robustness} & ECG-ID, BIDMC, CINC-2011, CINC-2014, teleECG and MIT-BIH arrhythmia & Compared AlexNet, VGG, ResNet, LSTM, GRU and BiLSTM for ECG noise detection.&  Limited generalization to unseen noise and lack of cross-dataset testing.\\
    \cite{satija2018review} & MITBIHA, PICC 2011, MIT-BIHSTC, Fantasia, MACE, MIHBIHNST, MIMIC-II, TELE & Automatic detection, localization, and classification of single and combined ECG noise & Signal quality assessment (SQA) methods are not robust for wearable ECG systems.\\
        \hline
        
    \end{tabular}
    }
\end{table*}

\section{Methodology}
\subsection{Datasets} \label{section: dataset}
We used four datasets in our analysis. One of these, the Activity Dataset (AD), was collected by us, while the other three—the Speech Performance Dataset (SD) \cite{sahu2024wearable}, the MIT-BIH NST dataset \cite{moody1984bih}, and the BUTQDB dataset \cite{nemcova2020brno}—are publicly available. However, SD was manually labeled by us, and we significantly extended the MIT-BIH NST dataset. All the datasets are described below.  

\noindent{\bf SD dataset (D1):} This dataset was collected during the speech activity. It is a 3-lead ECG dataset recorded using Shimmer sensors at 1024 Hz sampling rate, in a controlled lab setting. Participants were seated on a chair and given a topic to deliver a two-minute speech. They were allowed to do hand movements while seated. At a time, three participants were present in the room with a moderator, taking turns to give their speeches. A total of 97 healthy participants took part in the study, within the age range of 17–28 years with an average age of 21 years, consisting of 65 males and 32 females.

\noindent{\bf BUTQDB dataset (D2):} This large, publicly available dataset consists of 18 long-term single-lead ECG recordings collected from 15 healthy subjects (9 females, 6 males) aged between 21 and 83 years. The recordings were obtained while the subjects were engaged in ordinary everyday activities under real-world settings. Data was collected at a sampling frequency of 1,000 Hz, with a minimum recording length of 24 hours. This dataset was already labelled into three classes.
 (i) \textbf{Class 1}: Clean ECG signals.
 (ii) \textbf{Class 2}: Slightly noisy signals, but the {R-peaks} are still visible.
 (iii) \textbf{Class 3}: Completely noisy signals and no {R-peaks} are  visible. In this work, we considered Class 1 and Class 2 as clean since the R-peaks were clearly visible, while Class 3 was considered noisy.

\noindent {\bf MIT-BIH-NST dataset (D3):}
The original MIT-BIH-NST dataset is derived from the MIT-BIH Arrhythmia dataset~\cite{moody2001impact}. It comprises 12 half-hour, 2-lead ECG recordings collected at a sampling rate of 360 Hz and 3 half-hour noise recordings that represent common artifacts in ambulatory ECG signals. The noise recordings include baseline wander (BW), muscle artifacts (MA), and electrode motion artifacts (EM). The ECG recordings in MIT-BIH-NST were generated by introducing controlled noise into clean signals from the MIT-BIH Arrhythmia dataset using the nstdbgen script of the WFDB Python Package\footnote{\url{https://archive.physionet.org/physiotools/wfdb.shtml}}. Noise was added in alternating two-minute segments after the first five minutes of each recording, with predefined signal-to-noise ratios (SNRs). While the original  MIT-BIH-NST dataset applied noise to only two recordings (118 and 119), we extend the approach to the entire MIT-BIH Arrhythmia dataset. For this study, only MA and EM were considered, with SNR values of 0 dB and -6 dB. These specific SNR values were chosen because, at these levels, the R-peaks are no longer visible.

\noindent{\bf AD dataset (D4):}
This dataset was specifically collected to capture noisy ECG recordings during various real-world muscle activities. Data collection was conducted in a semi-controlled lab environment after obtaining informed consent under institutional ethics approval. Each 10-minute session included four conditions: 2 minutes of sitting, 2 minutes of standing with hand movements, 2 minutes of walking involving extensive muscle movement, and a 1-minute baseline period following each activity. ECG signals were recorded from seven participants (aged 20–25 years) using Shimmer sensors at a sampling rate of 1024 Hz, yielding approximately 70–80 minutes of total data. We publicly release D4 along with this paper \cite{ourdataset}.
  
\subsection{Data Annotation of SD and AD datasets}

Two undergraduate experienced ECG annotators labeled the SD and AD datasets using the open-source software Trainset\footnote{\url{https://trainset.geocene.com/}}. The annotators have  1.5 years of experience working specifically on ECG analysis. During the annotation process, the annotators first applied basic signal processing techniques and then identified the start and end indices of noisy segments (i.e., distorted QRS complexes) in each ECG signal. Based on these indices, Trainset automatically assigned a ``noisy'' label to the corresponding segment.

\begin{figure*}[h]
    \centering

    \begin{subfigure}[t]{0.16\textwidth}
        \centering  \includegraphics[width=\linewidth,height=0.10\textheight,keepaspectratio]{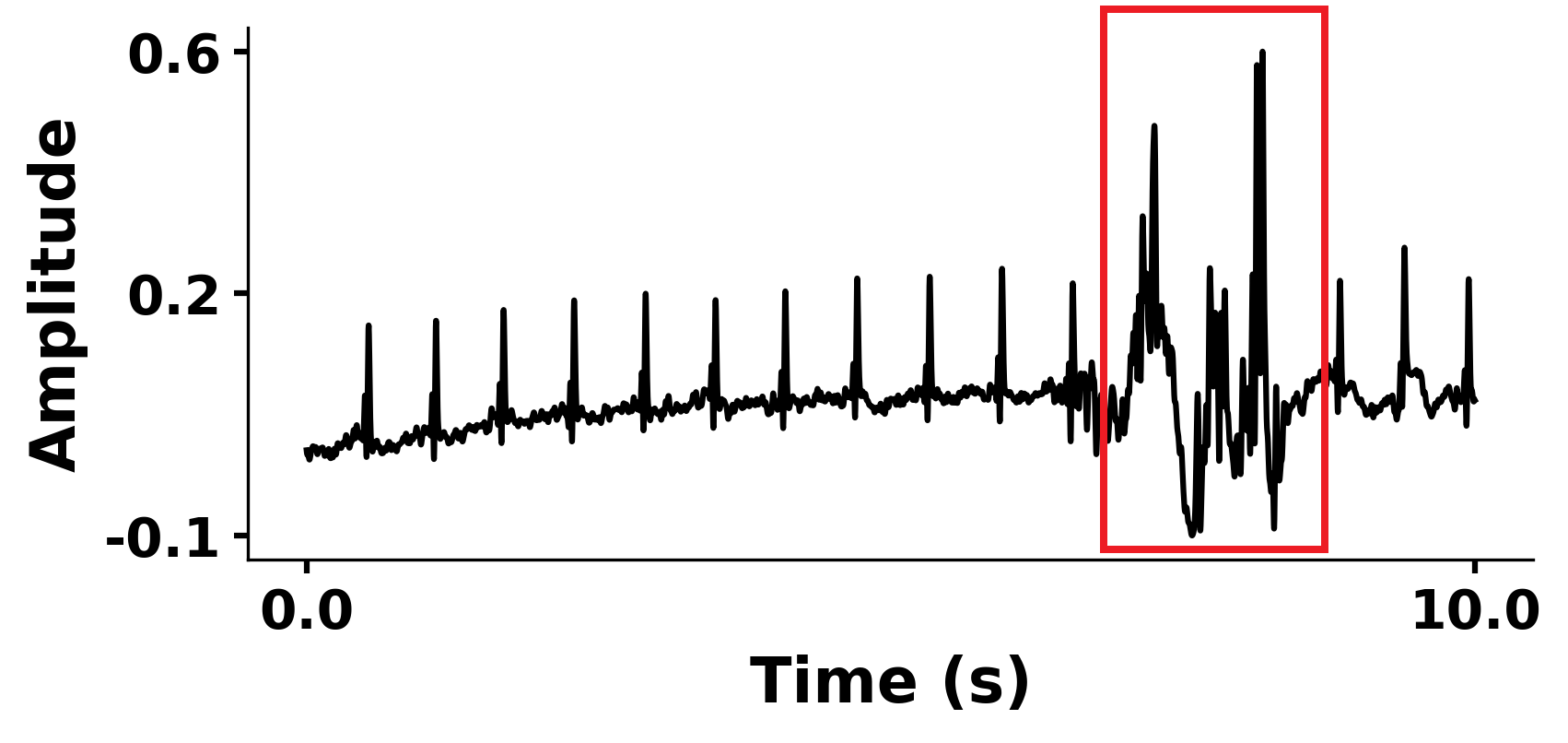}
        \caption{}
        \label{fig:d1a}
    \end{subfigure}\hfill
    \begin{subfigure}[t]{0.16\textwidth}
        \centering
        \includegraphics[width=\linewidth,height=0.10\textheight,keepaspectratio]{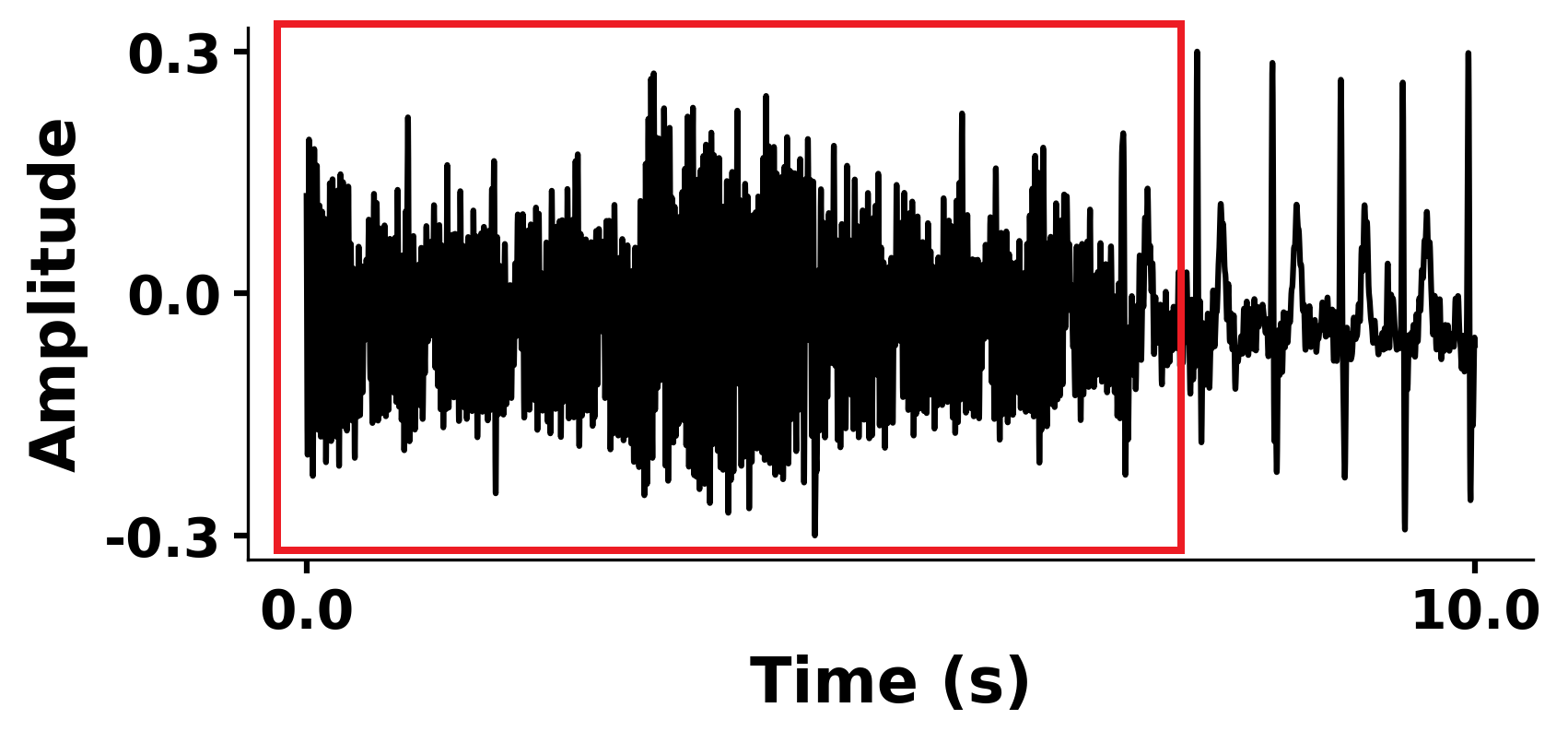}
        \caption{}
        \label{fig:d1b}
    \end{subfigure}\hfill
    \begin{subfigure}[t]{0.16\textwidth}
        \centering
        \includegraphics[width=\linewidth,height=0.10\textheight,keepaspectratio]{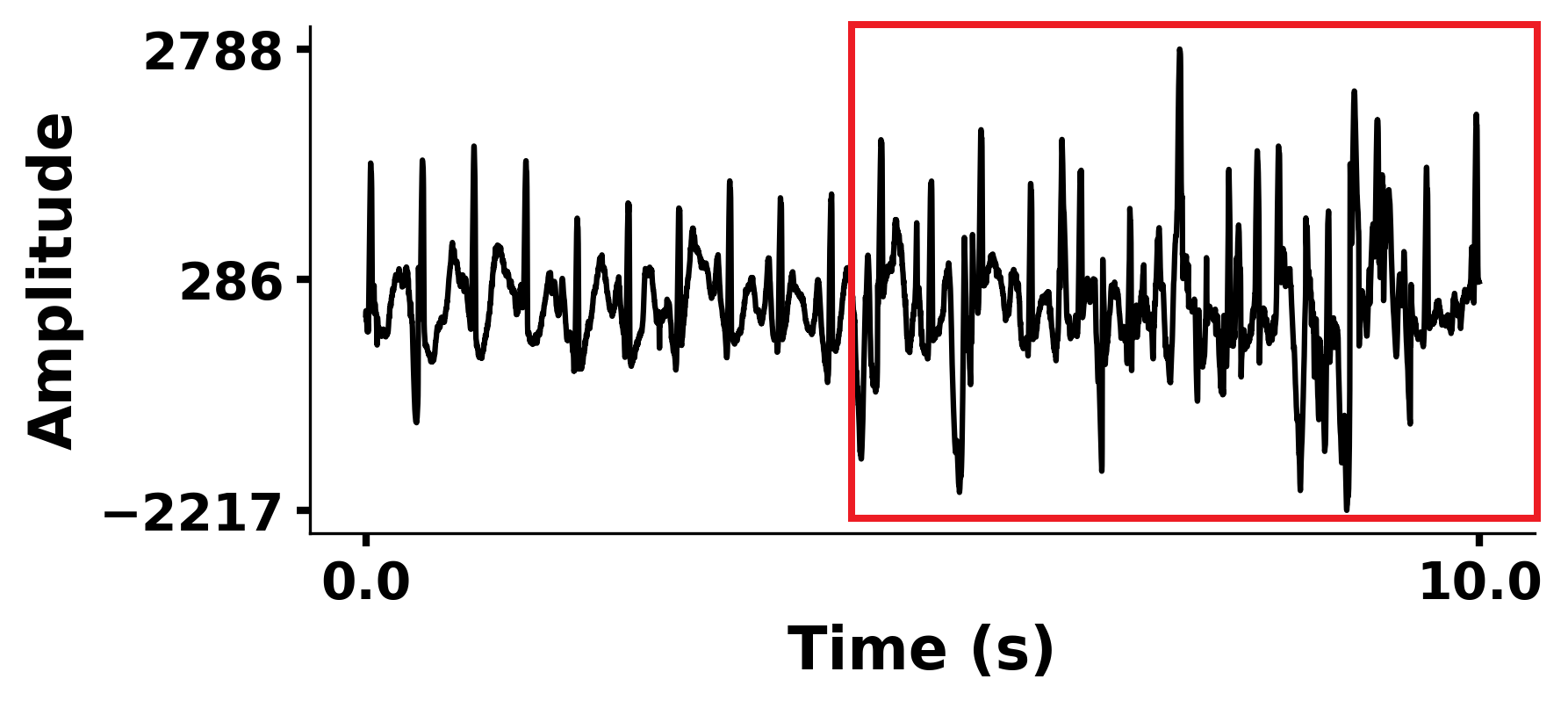}
        \caption{}
        \label{fig:d2a}
    \end{subfigure}\hfill
    \begin{subfigure}[t]{0.16\textwidth}
        \centering
        \includegraphics[width=\linewidth,height=0.10\textheight,keepaspectratio]{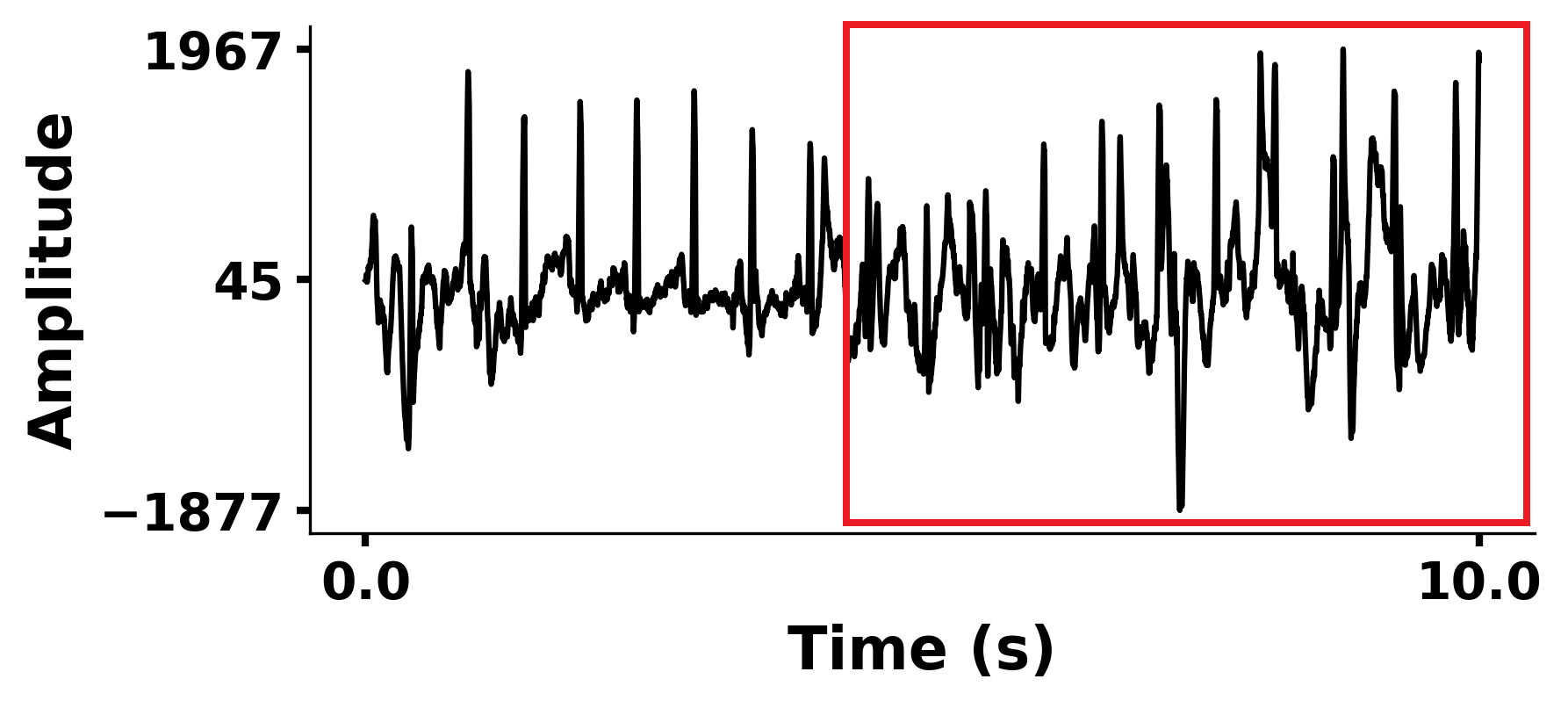}
        \caption{}
        \label{fig:d2b}
    \end{subfigure}\hfill
    \begin{subfigure}[t]{0.16\textwidth}
        \centering
        \includegraphics[width=\linewidth,height=0.10\textheight,keepaspectratio]{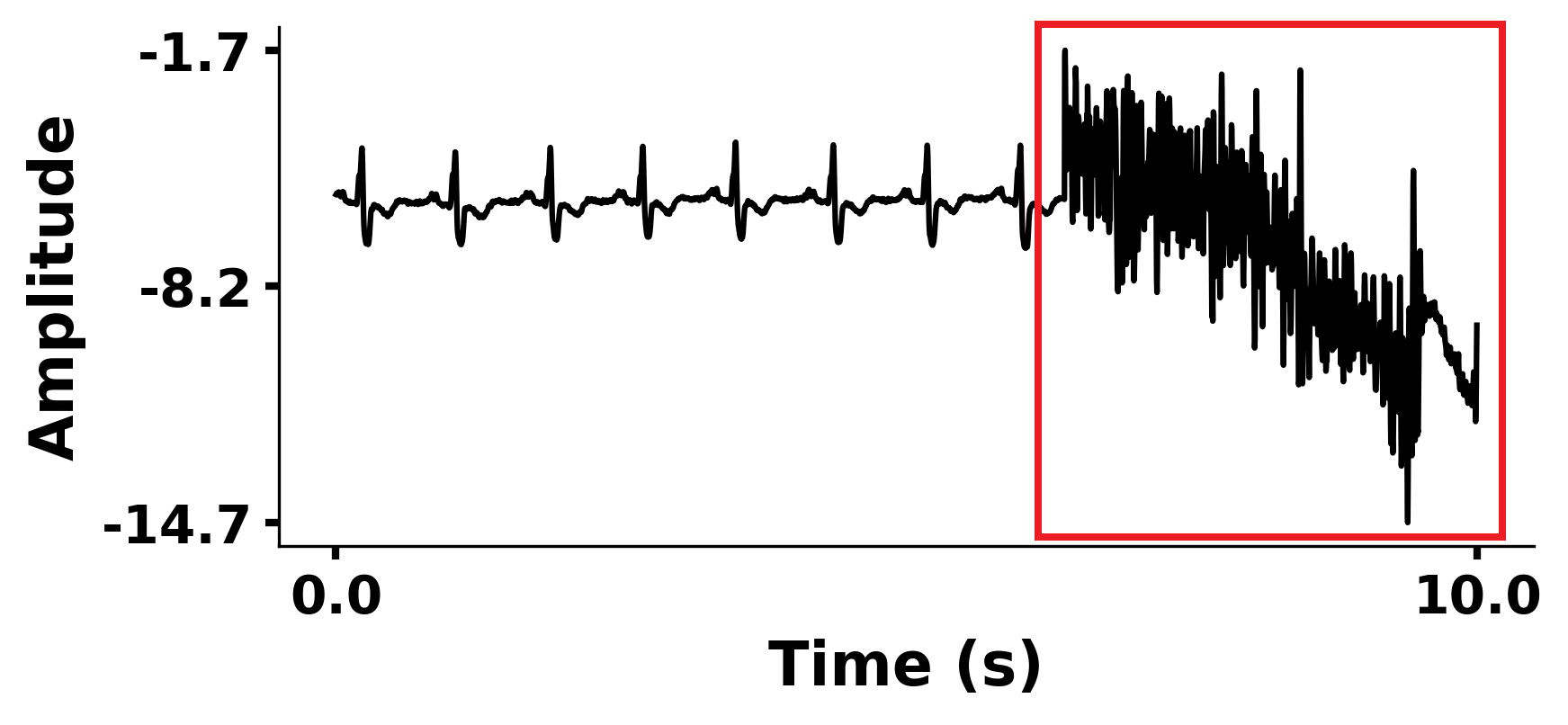}
        \caption{}
        \label{fig:d3a}
    \end{subfigure}\hfill
    \begin{subfigure}[t]{0.16\textwidth}
        \centering
        \includegraphics[width=\linewidth,height=0.10\textheight,keepaspectratio]{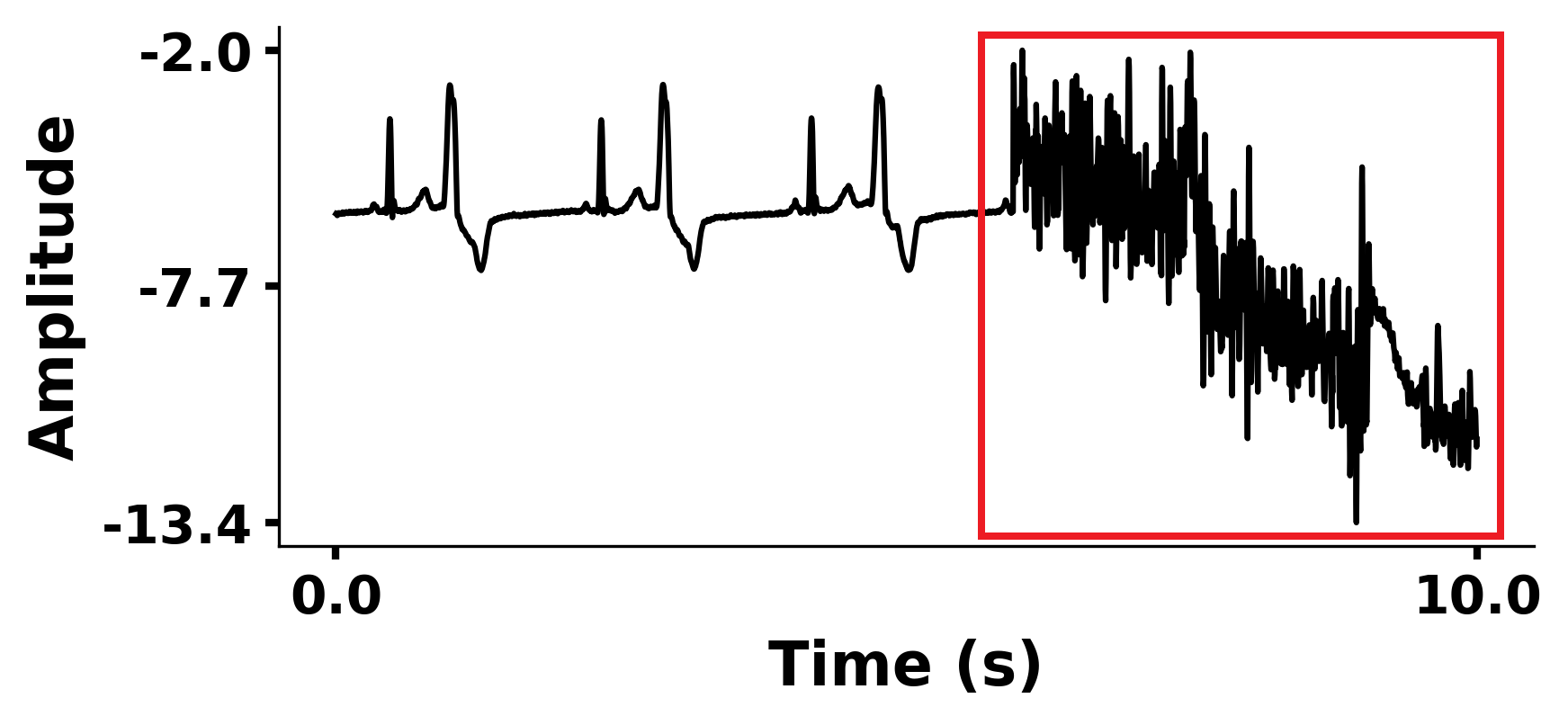}
        \caption{}
        \label{fig:d3b}
    \end{subfigure}

    \caption{Ten-second noisy ECG segments from three datasets. Subfigures (a) \& (b) correspond to D1; (c) \& (d) to D2; and (e) \& (f) to D3. Noise-contaminated regions are highlighted with red boxes.}
    \label{fig:final_grouped}
\end{figure*}

The annotators followed a single rule: \textit{``if R-peaks are not clearly visible, the segment from the last identifiable R-peak to the next clearly visible R-peak will be marked as noisy; otherwise, it will be labeled as clean.''} A label of 1 was assigned to noisy segments, while 0 was used for clean segments. In many cases, multiple noisy segments were present within a single ECG signal. Figure \ref{fig:final_grouped}(a),(b) illustrates the noisy segments identified by the annotators in one participant's data. Similarly, figures \ref{fig:final_grouped}(c),(d) and \ref{fig:final_grouped}(e),(f) show the noisy and clean segments in MIT-BIH-NST and BUTQDB, respectively. Throughout this paper, we refer to SD, BUTQDB, MIT-BIH-NST, and AD as \textbf{D1}, \textbf{D2}, \textbf{D3}, and \textbf{D4}, respectively. 

\subsection{ECG Segmentation and Filtering}

 To analyze changes at a granular level, we segmented ECG signals into 20 second windows with 50\% overlap. This window size was selected empirically based on performance across varying window lengths. Detailed sensitivity analysis is presented in Section~\ref{discussion}.

A 20-second window was labeled as noisy if at least 50\% of the indices within the window were noisy; otherwise, it was labeled as clean. It is important to note that while our datasets were originally annotated at the index level, we now assign labels at the segment (window) level. The number of clean and noisy segments in D1, D2, D3, and D4 is presented in Table \ref{tab:label_counts}.

After segmentation, we applied standard filtering techniques, including signal normalization, bandpass filtering, and a moving average filter, as part of the preprocessing step. We explored popular publicly available libraries  for physiological data processing, such as BioSPPy \cite{bota2024biosppy} and NeuroKit2 \cite{makowski2021neurokit2}. Among these, we found that the filtering parameters in BioSPPy were the most effective for cleaning ECG signals. It employs a bandpass Butterworth filter (3 Hz to 45 Hz), which effectively mitigates baseline wander, powerline interference, and minor muscle artifacts. 

\begin{table}[h!]
    \centering
    \caption{Number of clean \& noisy ECG segments in datasets.}
    \begin{tabular}{lccc}
        \toprule
        \textbf{Dataset} & \textbf{Clean (\#)} & \textbf{Noisy (\#)} & \textbf{Total (\#)} \\         \midrule
        D1& 973 & 163 & 1136 \\
        D2& 77395 & 2644 & 80039 \\
        D3& 19872 & 13064 & 32936 \\
        D4& 271 & 148 & 419 \\
        \bottomrule
    \end{tabular}
    \label{tab:label_counts}
\end{table}

\subsection{Feature Extraction}
We performed R-peak detection on the filtered ECG segments. We explored existing algorithms in the Neurokit2 \cite{makowski2021neurokit2} and BioSPPy \cite{bota2024biosppy} to identify an accurate R-peak detection algorithm. After going through the R-peak plots, we found that the Hamilton Segmenter \cite{hamilton1986quantitative} algorithm in BioSPPy performed best across all the datasets used, as it includes an additional step after peak detection, which corrects the detected R-peaks. This ensures that each R-peak is correctly positioned at the actual maximum point within a small window around the initially detected peak. Then we pass these R-peaks to the NeuroKit2 package, which then computes the RR interval (i.e., the time difference between consecutive R-peaks) and extract time-domain HRV features. We focused only on time-domain features because frequency-domain and non-linear HRV indices require larger ECG segments, which would undermine the purpose of noise detection, as noise artifacts are typically short. The computed time domain HRV features are shown in Table \ref{tab: computed_features}.

\begin{figure*}[!t]
    \centering
    \begin{subfigure}[b]{0.48\textwidth}
        \centering
    \includegraphics[width=\linewidth,height=3cm,keepaspectratio]{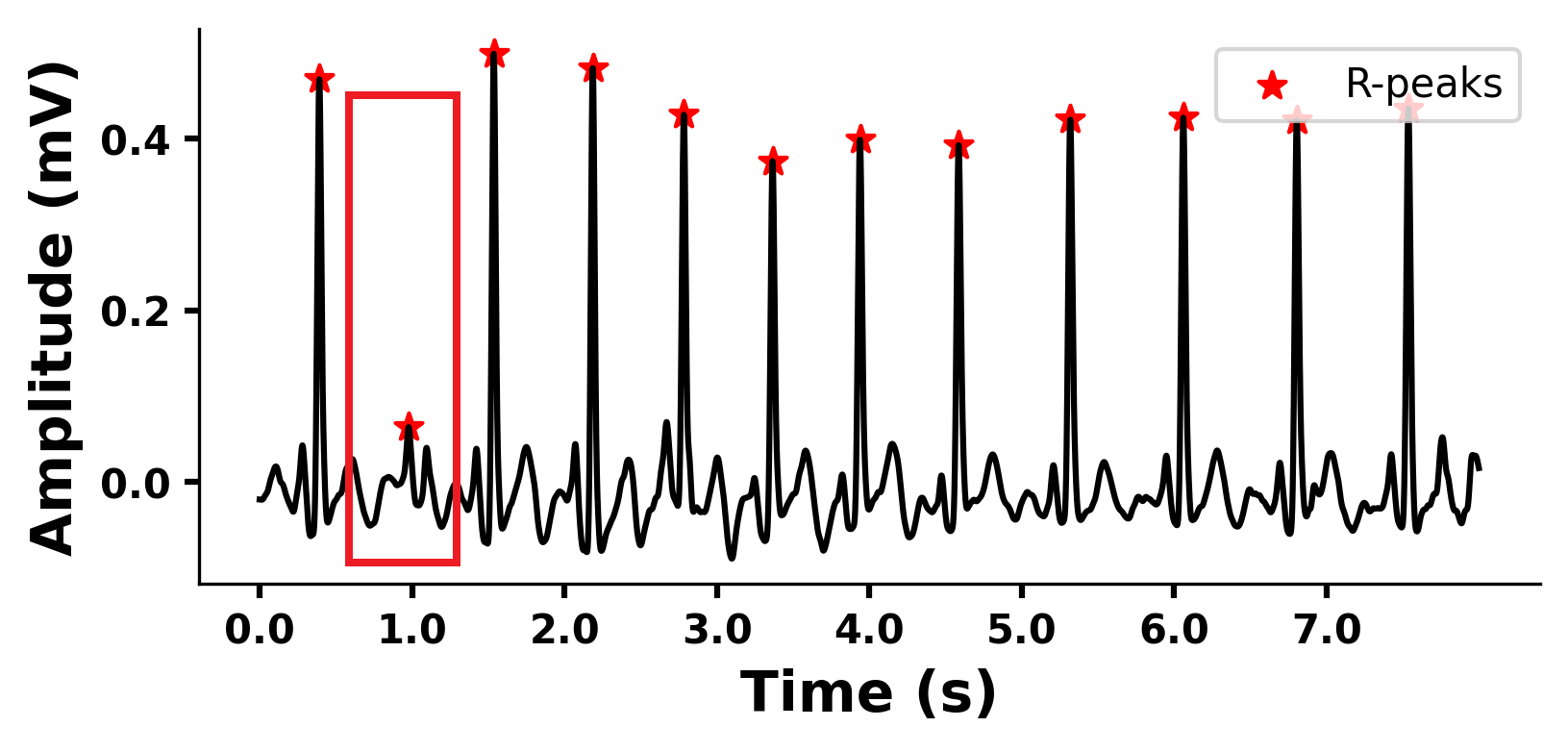}
        \caption{R-peaks with NeuroKit2}
        \label{fig:fig5a}
    \end{subfigure}
    \hfill
    \begin{subfigure}[b]{0.48\textwidth}
        \centering
        \includegraphics[width=\linewidth,height=3cm,keepaspectratio]{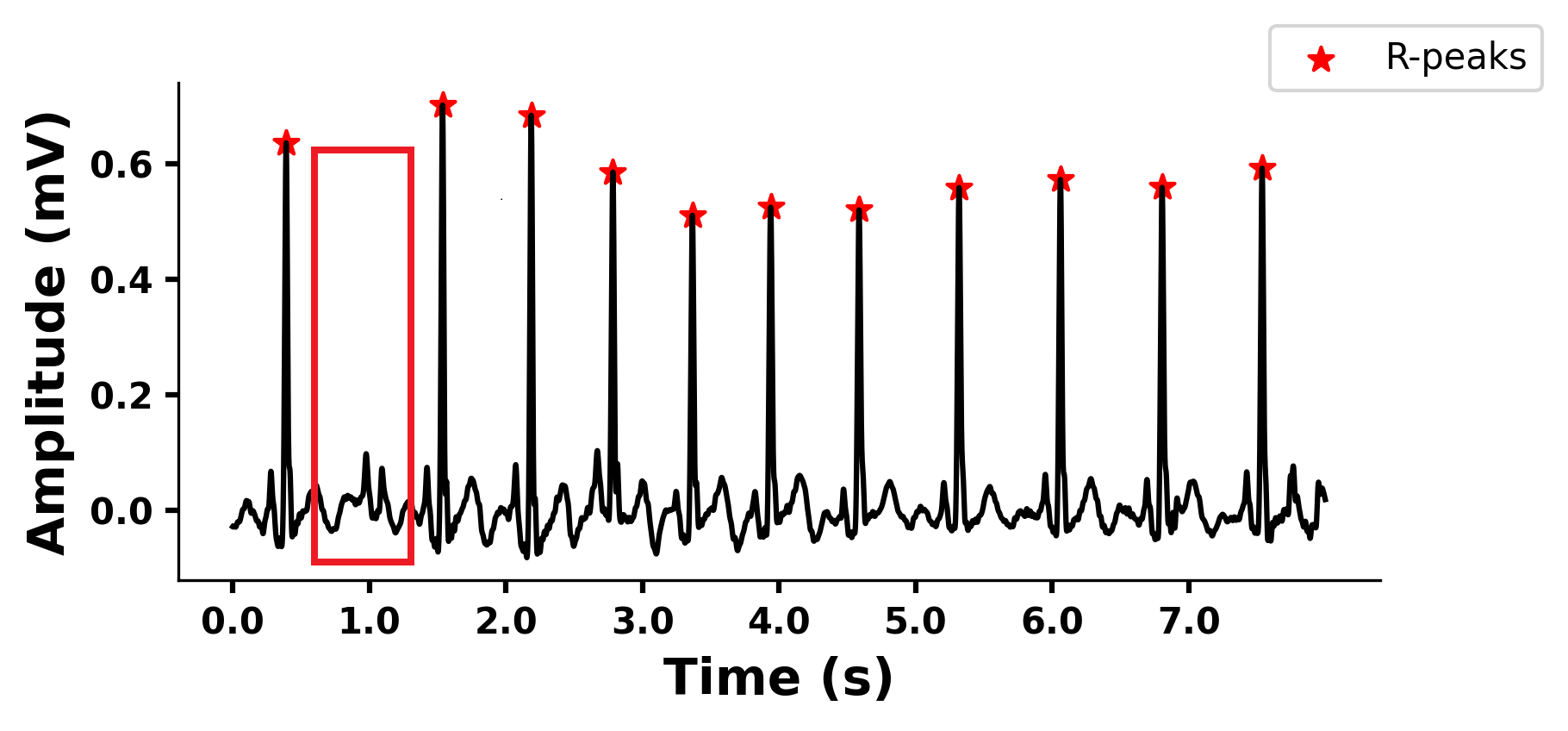}
        \caption{R-peaks with BioSPPy}
        \label{fig:fig5b}
    \end{subfigure}
    \caption{R-peaks detected with (a) Neurokit2 and (b) BioSPPy. The red boxes highlight regions where NeuroKit2 incorrectly identified a non-R-peak point, while BioSPPy correctly omitted it.}
    \label{fig:eR_peak_comparison}
\end{figure*}

\begin{table}[h!]
    \centering
    \scriptsize
    \caption{Computed Time-domain HRV Features}
    \setlength{\tabcolsep}{2.5pt} 
    \begin{tabular}{l l l}
    \toprule
    \textbf{Feature} & \textbf{Full Form} & \textbf{Description} \\
    \midrule
    MeanNN & Mean of NN intervals & Average time between heartbeats \\
    SDNN & Standard Deviation of NN intervals & Variability in heartbeat intervals \\
    RMSSD & Root Mean Square of Successive Differences & Short-term HRV measure \\
    SDSD & Standard Deviation of Successive Differences & Spread of successive RR intervals diff.\\
    CVNN & Coeff. of Variation of NN intervals & Relative variability of heartbeats \\
    CVSD & Coeff. of Variation of Successive Differences & Relative short-term variability \\
    MedianNN & Median of NN intervals & Median of RR intervals \\
    MadNN & Median Absolute Deviation of NN intervals & Robust measure of variability \\
    MCVNN & Median Coeff. of Variation of NN intervals & Median of the CV of NN intervals \\
    IQRNN & Interquartile Range of NN intervals & Spread of middle 50\% of RR intervals \\
    SDRMSSD & Standard Deviation of RMSSD & Variability in short-term HRV \\
    Prc20NN & 20th Percentile of NN intervals & Lower bound of 20\% of RR intervals \\
    Prc80NN & 80th Percentile of NN intervals & Upper bound of 80\% of RR intervals \\
    pNN50 & Percentage of NN intervals $>$ 50ms & Proportion of large RR interval diff. \\
    pNN20 & Percentage of NN intervals $>$ 20ms & Proportion of small RR interval diff. \\
    MinNN & Minimum NN interval & Shortest heartbeat interval \\
    MaxNN & Maximum NN interval & Longest heartbeat interval \\
    HTI & Heart Rate Turbulence Index & Heart rate response to premature beats \\
    TINN & Triangular Interpolation of NN intervals & Global HRV index from histogram width\\
    \bottomrule
    \end{tabular}
    \label{tab: computed_features}
\end{table}

\subsection{Classification Models}

We utilized Logistic Regression (LR), Random Forest (RF), Support Vector Machine (SVM), Decision Tree (DT), and Gradient Boosting (GB) for detection of noisy and clean ECG segments. LR was chosen for its simplicity in handling linearly separable data. RF was included for its ensemble learning capabilities, which improve robustness. SVM was selected for its ability to model complex decision boundaries. Additionally, we employed an Artificial Neural Network (ANN), 1D Convolutional Neural Network (CNN) and VGG~\cite{rahman2024robustness} to leverage deep learning techniques for classification.

\subsubsection{Evaluation Metrics}
To evaluate our trained classification models, we used Stratified 5-Fold Cross-Validation. In this method, the dataset is divided into five equal folds while preserving the class distribution in each fold, making it particularly suitable for our imbalanced datasets. Four folds are used for training, while the remaining fold is used for testing. This process is repeated for all combinations of folds to ensure a comprehensive evaluation. Next, we assessed models' performance using Accuracy, Precision, Recall, F1-score, and Area Under the Precision-Recall Curve (AUPRC). 

Accuracy measures the percentage of correctly classified samples as noisy or clean. Precision indicates the proportion of correctly classified noisy (or clean) samples out of all samples predicted as noisy (or clean).
Recall measures the proportion of actual noisy (or clean) samples that are correctly identified. 
The F1 score provides a harmonic mean of Precision and Recall, balancing both metrics. AUPRC summarizes the model’s Precision-Recall performance across different classification thresholds, evaluating its ability to distinguish between noisy and clean samples.

Due to the imbalanced nature of the datasets, we  used the weighted average method for accuracy, precision, recall, and F1-score to ensure a fair evaluation and prevent the majority class from dominating the results. Moreover, AUPRC is useful for understanding the classification performance in imbalanced datasets. 

\subsubsection{Parameter Settings}
The machine learning models were configured with their default hyperparameter in the scikit-learn library. LR used an L2 regularization penalty with a regularization strength parameter \( C = 1.0 \) and a maximum iteration limit of 100. The RF consisted of 100 decision trees, also using the Gini criterion. The SVM used a radial basis function (RBF) kernel with a regularization parameter \( C = 1.0 \). ANN model uses sequential dense layers: 64, 32, and 16 neurons with ReLU activations, ending with a sigmoid output for binary classification. It is compiled with Adam optimizer, binary cross-entropy loss, and tracks accuracy. The 1D CNN model uses two convolutional layers (64 and 32 filters, kernel size 3, ReLU, same padding) each followed by max pooling, then a flatten layer, a dense layer with 16 neurons, and a sigmoid output for binary classification. It is compiled with the Adam optimizer and binary cross-entropy loss. The VGG architecture was adopted from \cite{rahman2024robustness}.

\section{Results}

We present the classification results in two different settings: (i) Within-dataset classification - where models are trained and tested on the same dataset. (ii) Cross-dataset classification -  where models are trained on one dataset and tested on a different dataset.

\subsection{Within-dataset Classification}
Table \ref{tab:within_dataset_model_metrics} presents the  within-data classification results. We achieved the highest accuracy of 96.4\% and an AUPRC of 91.8\% on D1 (i.e., SD) using LR. Similar accuracy was observed with other classifiers; however, their AUPRC values were lower than that of LR. For D2 (i.e., BUTQDB), the highest accuracy of 99.7\% and an AUPRC of 98.4\% were achieved using RF. Similarly, for D3 (i.e., MIT-BIH-NST), RF achieved the highest accuracy of 93.6\% and an AUPRC of 97.6\%. LR performed well on D1 due to its simplicity and effectiveness, particularly with smaller datasets. The dataset D1 contains 1,136 data points with a class ratio of approximately 6:1, making it enough for LR to model effectively. However, LR struggles with larger and more complex datasets such as D2 and D3, which contain more than 80,000 and 32,000 data points respectively. Moreover, D2 has a class ratio of nearly 29:1, making it more challenging for LR to perform well. On closer inspection of results in D1, we found that RF achieved only 0.3\% lower accuracy and 0.2\% lower AUPRC than LR. Given the strong overall performance of the RF classifier within-dataset evaluation, we selected it for generalizability testing on the unseen datasets (i.e., cross-generalizability).

\setlength{\tabcolsep}{3.5pt}
\begin{table}[ht]
\centering
\small
\caption{Within-dataset classification results on datasets D1, D2, and D3 using classifiers - LR, RF, SVM, 1D CNN, ANN, and VGG.}
\begin{tabular}{l@{\hskip 0.2cm}l c c c c c}
\toprule
\textbf{Model} & \textbf{Dataset} & \textbf{Accuracy} & \textbf{Precision} & \textbf{Recall} & \textbf{F1 Score} & \textbf{AUPRC} \\ 
\midrule
 & \textbf{D1}  & \textbf{0.964} & \textbf{0.920} & \textbf{0.822} & \textbf{0.867} & \textbf{0.918} \\
LR & D2  & 0.995 & 0.917 & 0.947 & 0.931 & 0.968 \\
 & D3  & 0.912 & 0.876 & 0.906 & 0.891 & 0.944 \\
\midrule
 & D1  & 0.961 & 0.906 & 0.816 & 0.858 & 0.916 \\
RF & \textbf{D2}  & \textbf{0.997} & \textbf{0.943} & \textbf{0.955} & \textbf{0.949} & \textbf{0.984} \\
 & \textbf{D3}  & \textbf{0.936} & \textbf{0.911} & \textbf{0.929} & \textbf{0.920} & \textbf{0.976} \\
\midrule
 & D1  & 0.960 & 0.922 & 0.791 & 0.851 & 0.892 \\
SVM & D2  & 0.996 & 0.941 & 0.936 & 0.938 & 0.972 \\
 & D3  & 0.924 & 0.894 & 0.918 & 0.906 & 0.959 \\
\midrule
 & D1 & 0.954 & 0.955 & 0.954 & 0.954 & 0.902 \\
1D CNN & D2 & 0.996 & 0.996 & 0.996 & 0.996 & 0.983 \\
 & D3 & 0.931 & 0.931 & 0.931 & 0.931 & 0.968 \\
\midrule
 & D1 & 0.954 & 0.953 & 0.954 & 0.953 & 0.897 \\
ANN & D2 & 0.996 & 0.996 & 0.996 & 0.996 & 0.977 \\
 & D3 & 0.929 & 0.929 & 0.929 & 0.929 & 0.967 \\
 \midrule 
 & D1 & 0.8826 & 0.7790 & 0.8826 & 0.8275 & 0.1628 \\
 VGG & D2 & 0.9743 & 0.9501 & 0.9743 & 0.9620 & 0.2083 \\
 & D3 & 0.9174 & 0.8696 & 0.9174 & 0.8876 & 0.8790 \\
\bottomrule
\end{tabular}
\label{tab:within_dataset_model_metrics}
\end{table}

\subsection{Cross-dataset Classification}
 
\begin{figure}
    \centering
    \includegraphics[scale=0.35]{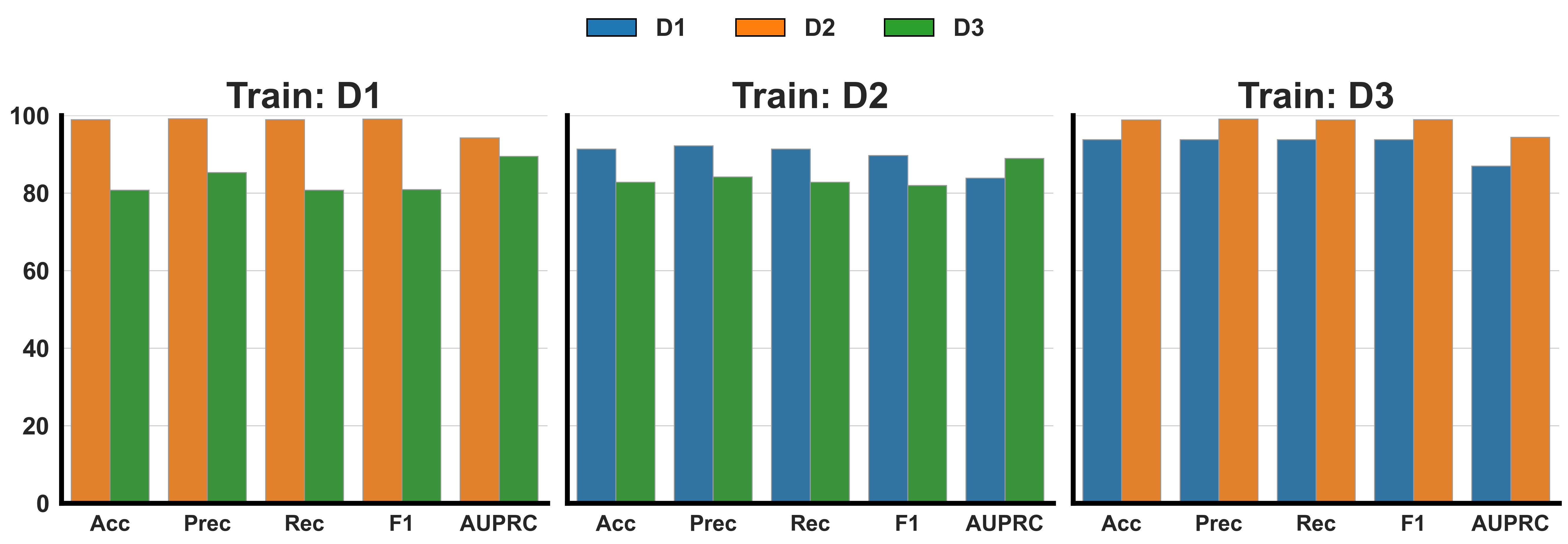}
 
    \caption{Cross-dataset classification results (Acc: Accuracy,  Prec: Precision, Rec: Recall, F1, AUPRC) using individual training datasets (D1, D2, and D3) with the RF classifier.}
    \label{fig:cross-dataset-results}
\end{figure}

Figure \ref{fig:cross-dataset-results} presents the results for training on one dataset and testing on other datasets. We found that the model trained on D1 performed better when tested on D2, achieving an accuracy of 99\% and an AUPRC of 94.3\%. However, when tested on D3, the accuracy dropped to 80.8\%. When trained on D2, the highest accuracy was 91.4\% on D1, but only 82.8\% on D3. In contrast, training on D3 resulted in the highest accuracy of 93.8\% and 98.9\% when tested on D1 and D2, respectively. The figure shows an interesting pattern: models trained and tested on real noisy datasets (D1 and D2) generalized well to each other but did not perform well on D3. However, training on D3 led to the best performance on both D1 and D2. This is because D3 contains a larger number of noisy samples, allowing the model to learn noisy patterns more effectively. 

\noindent {\bf Evaluation on combined training datasets:}
Table \ref{tab:cross_combined} presents the training results on combined datasets and testing on the remaining datasets. We found that training on D1 and D2 together and testing on D3 achieved an accuracy of 86.5\%, which was higher than training individually on D1 or D2 and testing on D3 (see Figure \ref{fig:cross-dataset-results}). This suggests that combining D1 and D2 helped the model learn the patterns of noisy and clean data more effectively. A similar trend was observed when training on combined D2 and D3 and testing on D1, where the accuracy improved compared to individual training and testing on D1. However, when training on combined D1 and D3 and testing on D2, the results were similar to individual training and testing on D2, achieving an accuracy of 99.2\%.

\begin{table}[ht]
\centering
\small
\caption{Cross-dataset classification results on using combined training datasets with the RF classifier.}
\begin{tabular}{l l c c c c c}
\toprule
\textbf{Train} & \textbf{Test} & \textbf{Accuracy} & \textbf{Precision} & \textbf{Recall} & \textbf{F1 Score} & \textbf{AUPRC} \\
\midrule
D2+D3 & D1 & 0.952 & 0.951 & 0.952 & 0.949 & 0.899 \\
D1+D3 & D2 & 0.992 & 0.993 & 0.992 & 0.992 & 0.943 \\
D1+D2 & D3 & 0.865 & 0.866 & 0.865 & 0.863 & 0.904 \\ \midrule
\multicolumn{2}{c}{\textbf{\textit{Average}}} & \textbf{0.936} & \textbf{0.937} & \textbf{0.936} & \textbf{0.935} & \textbf{0.915} \\
\bottomrule
\end{tabular}
\label{tab:cross_combined}
\end{table}

\begin{figure}[h!]
    \centering
    \includegraphics[scale=0.35]{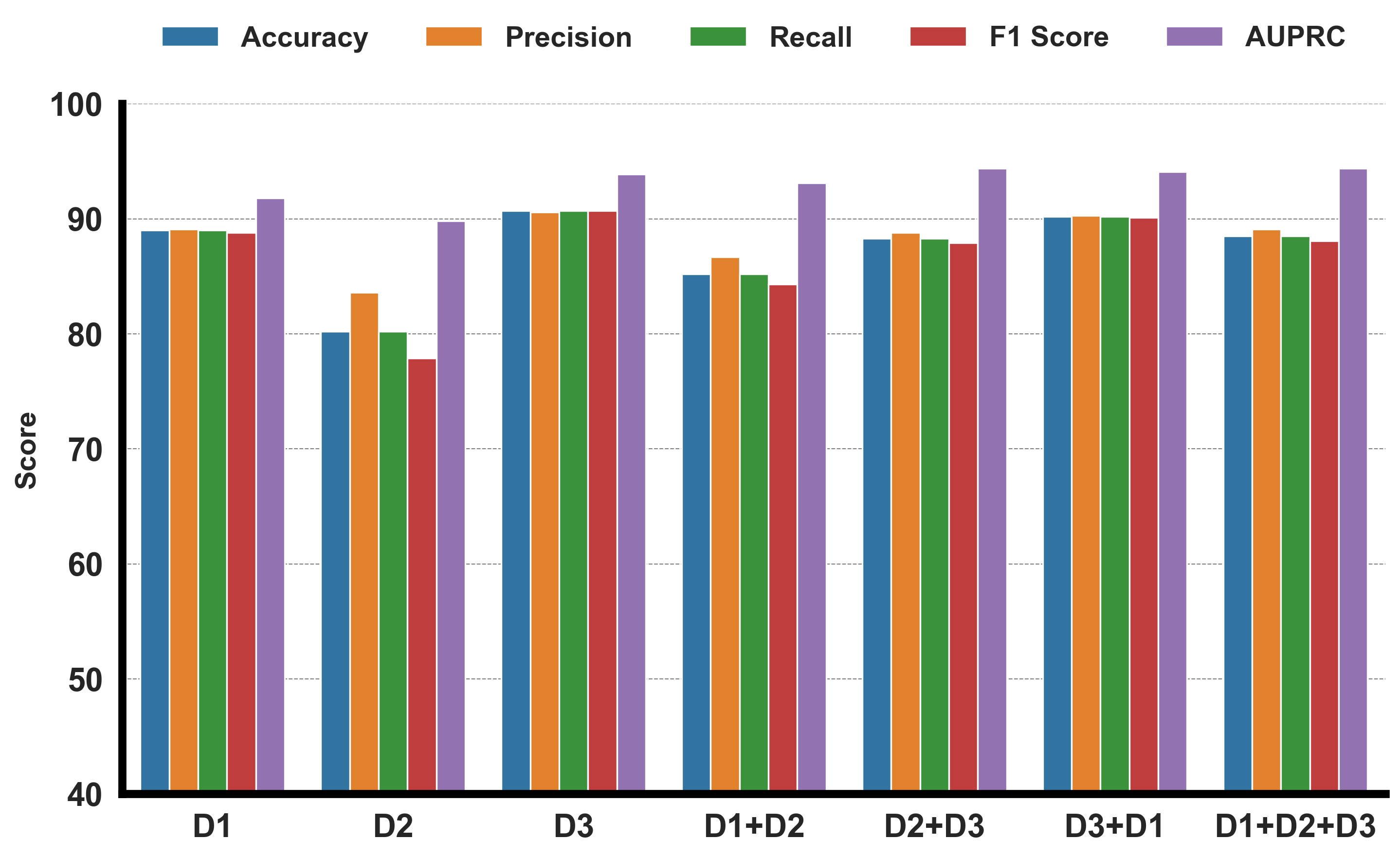}
    \caption{Cross-dataset classification results on the D4 dataset using the RF classifier.}
    \label{fig:Results_of_D4}
\end{figure}

\noindent{\bf Evaluation on special AD dataset:}
D4 (AD dataset) is relatively small compared to the other datasets; therefore, it was excluded from the training process. Instead, we trained the RF classifier using D1, D2, D3 and their various combinations, and evaluated its performance on the unseen D4 dataset. As shown in Figure \ref{fig:Results_of_D4}, the classifier generalized well to D4, achieving a highest accuracy of 90.7\% and an AUPRC of 94.4\%.

\section{Discussion and Limitations}\label{discussion}
\begin{figure}
    \centering
    \includegraphics[scale=0.3]{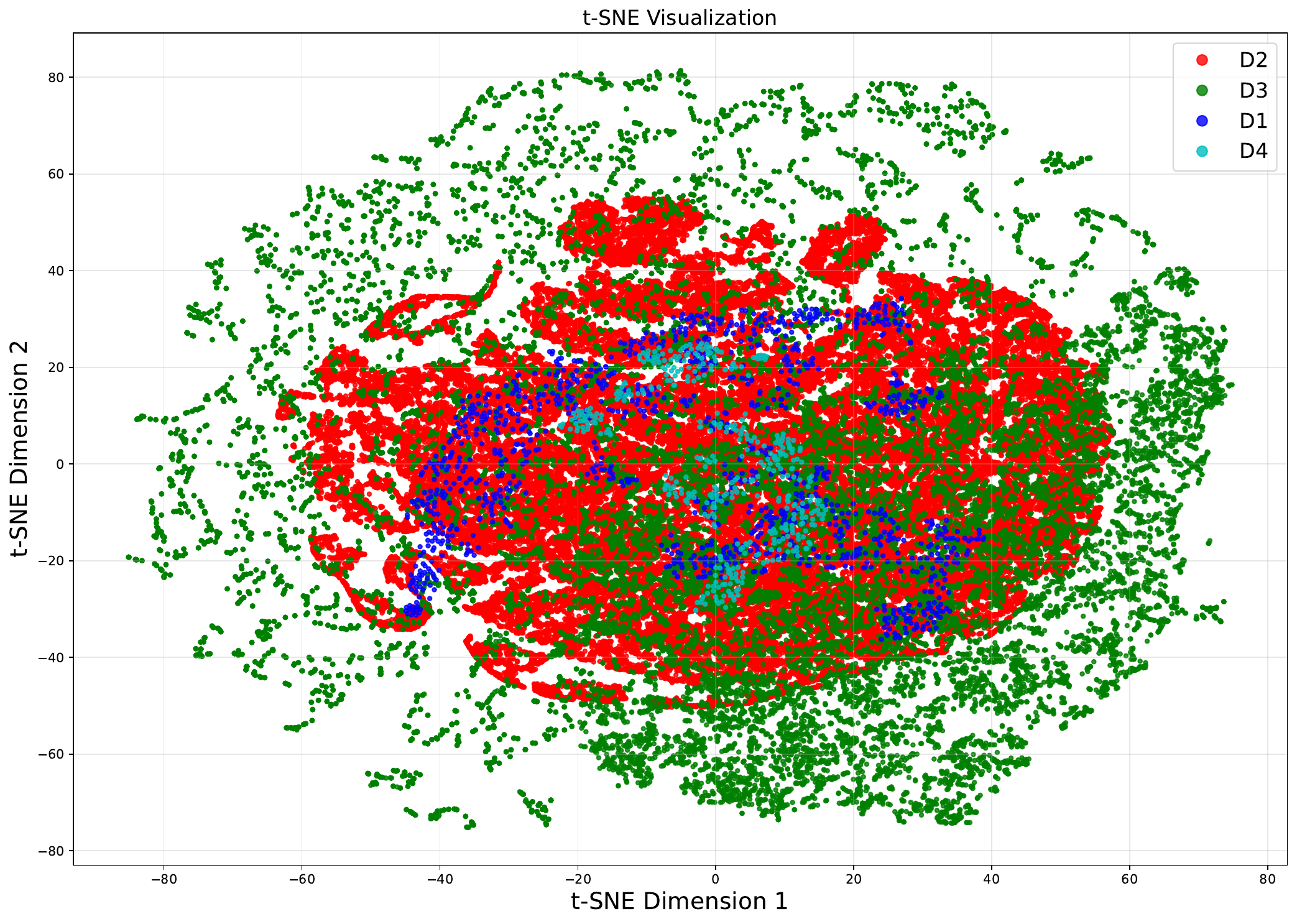}
    \caption{t-SNE plot of the datasets D1, D2, D3 and D4.}
    \label{fig:t-SNE_plot}
\end{figure}

The results from this study demonstrate that an HRV-based machine learning approach for ECG noise detection can be highly reliable, as shown by: 1) consistently high accuracy during individual training and testing, and 2) effective generalization, as observed in cross-dataset evaluation. In the cross-dataset evaluation, we note that datasets D1 and D2, which contain natural (non-induced) noise, generalize well when models are trained and tested on each other. However, when D1 and D2 are used for training and D3 for testing, the model does not perform. This is because D1 and D2 are highly imbalanced, with very few noisy labels compared to clean ones, resulting in limited variability in the noisy segments and an inability to properly classify the noisy labels in D3. Conversely, when the model is trained on the balanced D3 dataset and tested on D1 and D2, it yields very good results due to the broader spectrum of data across both labels. Moreover, the t-SNE plot in Figure~\ref{fig:t-SNE_plot} shows the relative variability order of the datasets, i.e., D1 $\subset$ D2 $\subset$ D3. This ordering provides additional insight into why models struggle to generalize to D3 in both within- and cross-dataset evaluations: the high variability in D3 introduces a wider range of noise patterns that are not well represented in the less diverse D1 and D2.

In Figure \ref{fig:Results_of_D4}, testing on D4 indicates that training on the highly imbalanced dataset D2 results in the worst accuracy, while training on D3 or the combined set D1+D3, which offer more balanced data, produces better results. This trend is consistent with our earlier cross-dataset evaluation, where models trained on more balanced datasets showed superior performance.
The performance drop likely stems from Random Forest’s tendency to favor the majority class due to bootstrap sampling and majority voting, reducing its effectiveness on imbalanced data.

\textit{Sensitivity analysis of window size:
As shown in Table~\ref{tab:rf_window_performance}, the 20-second window achieved the best overall performance across accuracy. Smaller windows lacked sufficient temporal context for accurate HRV calculations, while larger windows led to the loss of fine-grained noisy signatures. Thus, the 20-second window offers an optimal trade-off between data fidelity and robustness.}

\begin{table}[h!]
    \centering
    \caption{Performance metrics (\%) on D1 dataset across different window lengths using Random Forest Classifier.}
    \begin{tabular}{lccccc}
        \toprule
        \textbf{Window size} & \textbf{Accuracy} & \textbf{Precision} & \textbf{Recall} & \textbf{F1} & \textbf{AUPRC} \\
        \midrule
        05\_sec  & 87.9 & 84.7 & 65.1 & 73.5 & 79.1 \\
        10\_sec  & 86.4 & 84.8 & 71.7 & 77.6 & 86.2 \\
        15\_sec  & 86.2 & 84.6 & 78.6 & 81.3 & 88.9 \\
        \textbf{20\_sec}  & \textbf{96.1} & \textbf{91.8} & \textbf{80.3} & \textbf{85.6} & \textbf{89.6} \\
        25\_sec  & 82.8 & 83.7 & 77.8 & 80.6 & 89.3 \\
        30\_sec  & 81.4 & 82.7 & 79.1 & 80.6 & 90.7 \\
        \bottomrule
    \end{tabular}
    \label{tab:rf_window_performance}
\end{table}

\textit{Limitations:} In the current study, a fixed overlap window of 50\% was employed, which, while effective, results in the generation of a large number of instances and consequently increases the overall processing time. Our method relies on accurately detecting R peaks in order to compute HRV features. If R peaks are misclassified, the resulting feature values become distorted and can lead to incorrect noise detection. Future work could explore the use of a dynamic, adaptable overlap window that adjusts based on the signal characteristics, potentially optimizing both computational efficiency and detection performance.

\section{Conclusion}
We proposed an automated noise detection approach for ECG signals using HRV features. To assess its generalizability, we conducted extensive experiments across four datasets collected under varying conditions and using different wearable devices. By training machine learning models to classify ECG segments as noisy or clean based on HRV patterns, we demonstrated that our approach consistently achieves high accuracy across diverse data sources.

These results highlight the robustness and generalizability of our method, underscoring its potential for real-world deployment in applications such as real-time ECG monitoring and AI-driven cardiovascular or mental health assessment. Our work also contributes a curated, labeled ECG dataset to facilitate further research in this area.

\bibliographystyle{acm}
\balance
\scriptsize
\bibliography{citations}

\end{document}